\documentclass[letterpaper]{article} 
\usepackage[submission]{aaai25}  
\usepackage{times}  
\usepackage{helvet}  
\usepackage{courier}  
\usepackage{kotex}
\usepackage{booktabs}
\usepackage{array}  
\usepackage{tabularx}
\usepackage{listings}
\usepackage{amsmath}
\usepackage[hyphens]{url}  
\usepackage{graphicx} 
\usepackage{natbib}  
\usepackage{caption} 
\usepackage{algorithm}
\usepackage{algorithmic}
\usepackage{amsmath}
\usepackage{booktabs}
\usepackage{listings}

\usepackage{newfloat}
\usepackage{listings}
\DeclareCaptionStyle{ruled}{labelfont=normalfont,labelsep=colon,strut=off} 
\floatstyle{ruled}
\newfloat{listing}{tb}{lst}{}
\floatname{listing}{Listing}

\title{MathBridge: A Large Corpus Dataset for Translating Spoken Mathematical Expressions into \LaTeX{} Formulas for Improved Readability}
\author{
    Kyudan Jung\textsuperscript{\rm 1}\equalcontrib,
    Sieun Hyeon\textsuperscript{\rm 2}\equalcontrib,
    Jeong Youn Kwon\textsuperscript{\rm 1},
    Nam-Joon Kim\textsuperscript{\rm 2},
    Hyun Gon Ryu\textsuperscript{\rm 3},
    Hyuk-Jae Lee\textsuperscript{\rm 2},
    Jaeyoung Do\textsuperscript{\rm 2}
}
\affiliations{
    \textsuperscript{\rm 1}Chung-Ang University\\
    \textsuperscript{\rm 2}Seoul National University\\
    \textsuperscript{\rm 3}NVIDIA\\

}

\usepackage{bibentry}

\begin{document}

\maketitle

\begin{abstract}




Improving the readability of mathematical expressions in text-based document such as subtitle of mathematical video, is an significant task.
To achieve this, mathematical expressions should be convert to compiled formulas.
For instance, the spoken expression ``x equals minus b plus or minus the square root of b squared minus four a c, all over two a'' from automatic speech recognition is more readily comprehensible when displayed as a compiled formula $x = \frac{-b \pm \sqrt{b^2 - 4ac}}{2a}$.
To convert mathematical spoken sentences to compiled formulas, two processes are required: spoken sentences are converted into LaTeX formulas, and LaTeX formulas are converted into compiled formulas.
The latter can be managed by using LaTeX engines.
However, there is no way to do the former effectively.
Even if we try to solve this using language models, there is no paired data between spoken sentences and LaTeX formulas to train it. In this paper, we introduce MathBridge, the first extensive dataset for translating mathematical spoken sentences into LaTeX formulas.
MathBridge comprises approximately 23 million LaTeX formulas paired with the corresponding mathematical spoken sentences.
Through comprehensive evaluations, including fine-tuning with proposed data, we discovered that MathBridge significantly enhances the capabilities of pretrained language models for converting to LaTeX formulas from mathematical spoken sentences.
Specifically, for the T5-large model, the sacreBLEU score increased from 4.77 to 46.8, demonstrating substantial enhancement. 
\end{abstract}

\begin{figure*}[t]
    \centering
    \includegraphics[width=1.\textwidth, trim=14mm 80mm 38mm 80mm, clip]{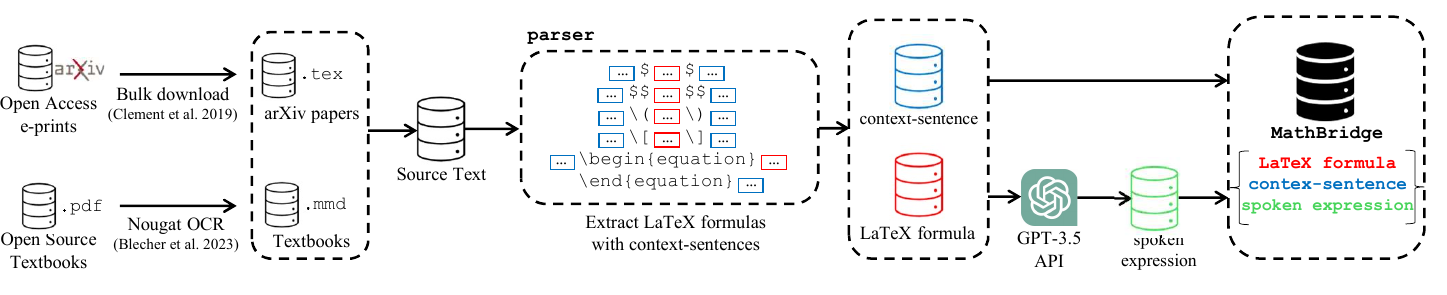}
    \caption{Illustration of process to construct MathBridge.}
    \label{fig:enter-label}
\end{figure*}

\section{Introduction}
Many students access open-source courses through various media sources. These courses often cover subjects such as natural sciences, engineering, artificial intelligence, and astronomy, where mathematical formulas are frequently discussed.Students with language barriers and students who are hearing impaired rely on subtitles to comprehend the lecturer's spoken content. When it comes to mathematical formulas, they must understand them through English text from ASR that vocalizes the notation displayed in the subtitles.

Unlike general English, interpreting formulas through vocalized English rather than as compiled formulas can be cognitively challenging. This is particularly true for complex formulas in which the difference becomes pronounced. As shown in Table 1, viewing the formula as a compiled format is cognitively easier than interpreting it through its spoken English equivalent, affecting both the speed of comprehension and perceived difficulty.

\begin{table}[t]
    \centering
    \setlength{\tabcolsep}{1pt}
    \footnotesize
    \begin{tabular}{>{\raggedright\arraybackslash}p{3cm} >{\raggedright\arraybackslash}p{5.2cm}}
    \toprule
        \textbf{Spoken Sentence} & x equals minus b plus or minus the square root of b squared minus four a c, all over two a \\
    \midrule
        \textbf{LaTeX Formula} & \$x = \textbackslash frac\{-b \textbackslash pm \textbackslash sqrt \{b \textasciicircum  2 - 4ac\}\}\{2a\}\$\\

    \midrule
        \textbf{Compiled Formula} &  { $ x = \frac{-b \pm \sqrt{b^2 - 4ac}}{2a} $} \\

    \bottomrule
    \end{tabular}
    \caption{Examples of mathematical spoken sentence, its corresponding LaTeX formula, and the compiled formula.}
    \label{tab:my_label}
\end{table}

To address this issue, the process of converting a spoken sentence into a compiled formula, which we will refer to as the ``text-to-formula'' process, can be broken down into two stages.The first stage involves converting the spoken sentence into a LaTeX format (text-to-LaTeX).The second stage requires compiling LaTeX into a compiled formula (LaTeX-to-formula), which is already possible with various compilers.\footnote{www.mathjax.org}However, the first stage poses a challenge. The vocalized spoken sentence of the formula does not have a one-to-one correspondence with the LaTeX syntax. Moreover, the extensive variety of TeX syntax makes it impractical to create a dictionary-style conversion for every possible grammar, as this approach would be too exhaustive.

We devised a solution for the text-to-LaTeX task using a language model.This involves developing a language model that translates mathematical expressions from ASR into LaTeX format.The LaTeX output from this model can then be inputted into a LaTeX engine to generate formulas that are visually similar to those in Table 1. However, a significant challenge remains, as there is currently no existing dataset of paired LaTeX formulas and their corresponding English vocalizations that are necessary for supervised learning. Although Quiniou created a dataset containing mathematical vocalizations, it was limited to French vocalizations, which restricts its applicability \cite{2-a}. 

Consequently, we developed MathBridge, a dataset aimed at translating Mathematical English into LaTeX format. MathBridge facilitates the conversion of expressions into LaTeX, which can subsequently be converted into compiled formulas. This enhances the accessibility of mathematical formulas and improves students' comprehension of online mathematics courses. LaTeX formulas were sourced from pre-prints and articles published on arXiv\footnote{https://arxiv.org} in 2023 and open-source textbooks. Open-source textbooks include materials from middle school, high school, and university courses in subjects such as calculus and algebra.From this extensive source data, we extracted LaTeX formulas as well as the sentences immediately preceding and following each formula, such as "The equation is" or "which is important," which we refer to as `context\_before' and `context\_after', respectively.When referring to both collectively, we will use the term `context-sentence.’We used the GPT-3.5 API to obtain high-quality sentence data, where each sentence corresponds to the spoken version of a LaTeX formula.We carefully selected valid formulas and filtered out noisy sentences, resulting in 23 million pairs. 

In our experiments with pretrained language models (PLMs), we utilized MathBridge to fine-tune the PLMs into a text-to-LaTeX translator.This establishes a robust baseline for translating spoken expressions into the LaTeX format in future research. We trained the model by inputting both formulas and context-sentences. Moreover, our analysis of the experimental results revealed that a single spoken mathematical expression can correspond to several different LaTeX formulas. Furthermore, we point out that the conventional evaluation metric is not suitable for assessing LaTeX. Subsequently, we propose three conditions that an ideal metric should satisfy. In summary, our contributions are as follows:
\begin{itemize}
\item 
We present MathBridge, the first large-scale text dataset for translating mathematical English into the LaTeX format. 
\item Experiments on MathBridge demonstrate that MathBridge is an excellent dataset for effectively converting mathematical expressions into LaTeX format.
\item We have determined that conventional criteria such as BLEU, ROUGE, WER, and CER are inappropriate for evaluating LaTeX text alignment. We also proposed conditions that an appropriate metric for evaluating LaTeX should satisfy. \end{itemize}

The dataset and fine-tuned models are publicly available.\footnote{https://github.com/MathBridge}

\section{Related Work}
Previous studies on \textbf{multilingual translation} have primarily focused on natural languages \cite{4-a,4-b,4-d,4-e,4-f}. These efforts aimed to convert English into various national languages rather than artificial languages. The advent of pre-trained language models, such as BERT \cite{BERT} and GPT \cite{GPT} has revolutionized the field of natural language processing, including translation. These models, pretrained on an extensive corpus, can be fine-tuned for specific translation tasks to achieve state-of-the-art results.

Broadly speaking,\textbf{ translating SQL queries and natural languages } can also be regarded as a form of translation. SQL, like LaTeX, is a computer language governed by a set of defined rules that present similarities in structured linguistic conversion. Research in this area includes various studies of SQL-to-text and text-to-SQL translations \cite{5-a,5-b,5-c,5-d}.

Only a few studies have explored the relationship between spoken English and LaTeX. Further research in this field could significantly improve educational technologies, especially for non-native English speakers and individuals with hearing impairments, by enhancing their access to scientific and mathematical content.

\section{Dataset Construction}



\subsection{Source Text}
To gather LaTeX formulas and their corresponding English vocalization sentences, we initially focused on collecting LaTeX formulas because they are easier to obtain than spoken English sentence data. We extracted LaTeX formula data from papers uploaded to arXiv and open-source textbooks.

\subsubsection{arXiv papers}
On the arXiv website, it is possible to view papers as HTML files through a browser without directly downloading LaTeX files. Although it is feasible to extract LaTeX formulas using HTML tags that represent formulas, this method is not recommended because it can place undue stress on arXiv servers. Instead, as shown in Figure 1, we followed the method proposed by Clement to bulk download the arXiv paper files \cite{1-a}.  This method utilizes a manifest file that lists the metadata of papers, allowing the downloading of PDF or source files via AWS servers. Because we did not require PDF files or image files from the source files, we downloaded only. tex extension files. Using metadata information, we downloaded the files uploaded in 2023. This approach not only extracts mathematical formulas but also enables extraction from all categories of papers provided by arXiv. Thus, the extracted formulas span a diverse range of disciplines. The distribution of papers across different fields is illustrated in Figure 2(a), where `etc.' includes fields such as economics, quantitative biology, and statistics.
\begin{figure}
    \centering
    \includegraphics[width=0.9\linewidth, trim=5mm 5mm 15mm 5mm, clip]{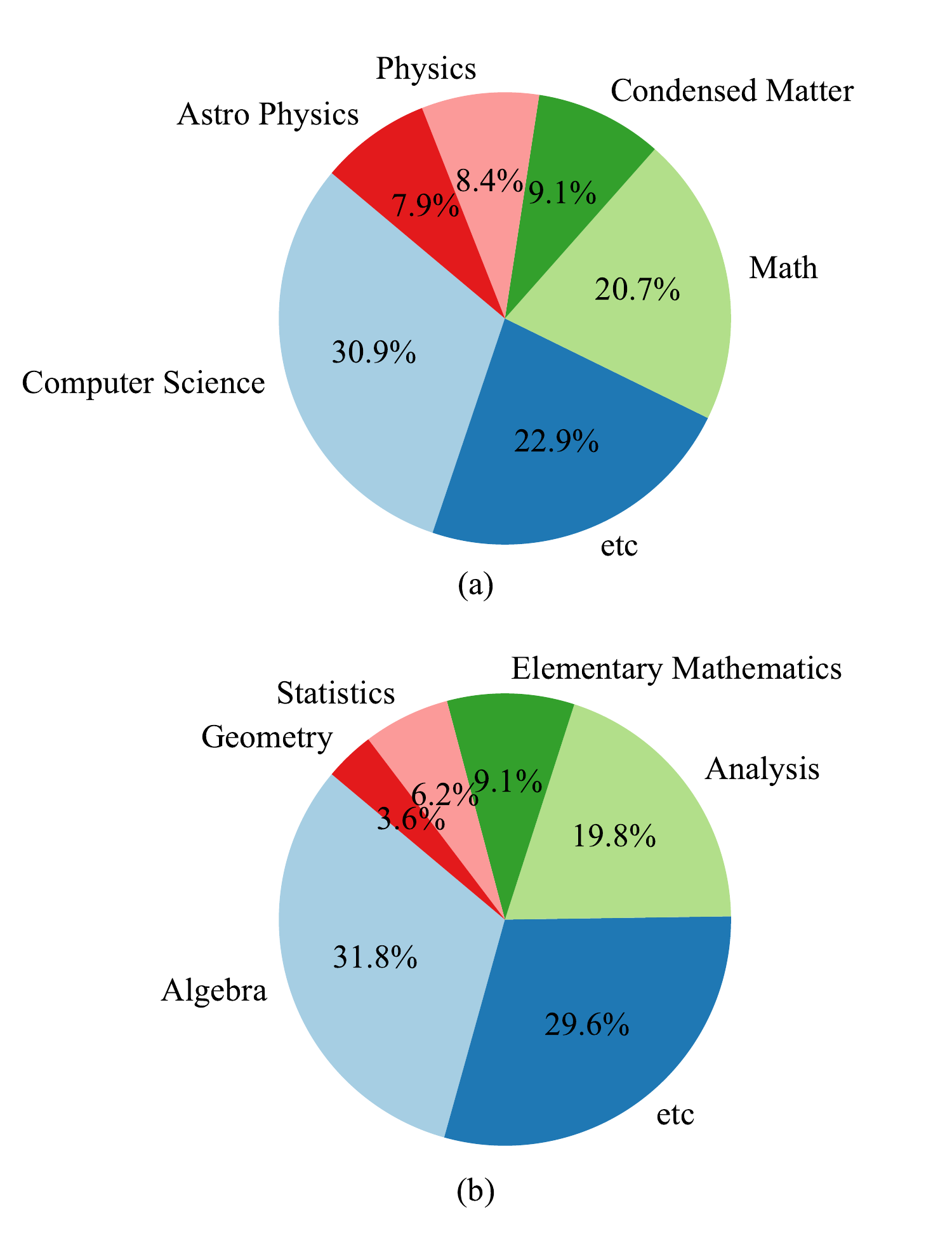}
    \caption{(a) Proportion of textbook fields in the source text. (b) Proportion of arXiv papers' fields in the source text.}
    \label{fig:enter-label}
\end{figure}

\subsubsection{textbook}
We determined that the formulas available for arXiv tended to be of higher complexity and were suitable mainly for advanced studies. Consequently, we decided to include formulas from elementary, middle, high school, and university mathematics textbooks that are available as open sources. We initially collected the PDFs of open-source textbooks. \footnote{https://bestedlessons.org/high-school/}\footnote{https://textbooks.aimath.org/textbooks/approved-textbooks/} These textbooks range from Elementary Mathematics, which contains content suitable for younger students, to more advanced topics such as Analysis and Algebra. The collected data are in PDF format, which does not allow for the direct extraction of formulas. Therefore, we used the optical character recognition (OCR) tool nougat \cite{3-a_nougat} to process the PDF files. A significant advantage of using Nougat is that it provides OCR results in multi-markdown (mmd) files that include LaTeX. We compiled a collection of mathematics textbook data in mmd format. The distribution of textbook pages by subject area is shown in Figure 2(b), where `etc.' includes fields such as Data Science, Machine Learning, and Discrete Mathematics.

\begin{table*}[ht]
\centering
\footnotesize
\setlength{\tabcolsep}{1.5pt}
\begin{tabular*}{\textwidth}{@{\extracolsep{\fill}}>{\raggedleft\arraybackslash}p{35mm} >{\centering\arraybackslash}p{4cm} p{3.5cm} p{4.5cm}}

\toprule
\textbf{context\_before} & \textbf{LaTeX formula} & \textbf{context\_after} & \textbf{spoken\_sentence} \\
\midrule
and we adopted the convention that &  \texttt{\$x\_0=x(t)\$} & so that the index corresponding to the current physical time & x sub zero equals x of t.\\
\midrule
thereby showing that &  \texttt{\$ \textbackslash phi\_\{H\_k\}\text{(}y\text{)}=E\text{(}y\text{)}\$} & is a linear mapping. & Phi sub H sub k of y equals E of y. \\
\midrule
the cluster satisfying & \texttt{\$||x\_i-x\_j||\textasciicircum2\_2\>\textbackslash alpha\$} & if sample & Norm of the difference between x sub i and x sub j, squared, subscript two, is greater than alpha.	\\
\midrule
Note that &  \texttt{\$\text{E}[\text{e\_t\textasciicircum 2}]\text{=}\text{E}[\text{sigma\_t\textasciicircum 2}]\text{=}\text{1}\$} & The endogenous variable is	& Expectation of e sub t squared equals expectation of sigma sub t squared equals one. \\
\bottomrule
\end{tabular*}

\caption{Examples of MathBridge dataset.}
\end{table*}
\subsection{\LaTeX{} Formula}
In this section, we discuss how we extracted LaTeX formulas and context-sentence. The example data for MathBridge is shown in Table 2.
The first algorithm applied during the extraction process involves replacing custom commands in the source text dataset with the standard LaTeX syntax. The authors of arXiv papers often define new custom commands at the beginning of their work. tex files using commands such as \textbackslash def, \textbackslash newcommand, and \textbackslash renewcommand. These custom commands frequently do not conform to the standard LaTeX syntax.
Without addressing this, extracting LaTeX would also extract these custom commands, leading to errors during the compilation and incorrect compiling of formula.
To resolve this issue, we replaced all custom commands with their original LaTeX syntax equivalents.

Next, we defined a parser capable of extracting formulas. The parsing conditions were specified as follows: 1) between \$ and \$, 2) between \$\$ and \$\$, 3) between \textbackslash ( and \textbackslash ), 4) between \textbackslash [ and \textbackslash ], 5) between \textbackslash begin\{equation\} and \textbackslash end\{equation\}. During the extraction process, we chose not to use the regular expression method. This decision was made because symbols \$ and \$\$ were used to both start and end the formula. Regular expressions would mistakenly recognize the closing symbol as an opening symbol because of their identical nature. To overcome this issue, our code was structured to sequentially search for each other. tex file from the beginning, storing formulas between each pair of \$ or \$\$ symbols encountered. This method proved to be more accurate for extracting formulas than regular expressions. Using this parser, we extracted approximately 48 million formulas from arXiv papers and 1 million formulas from textbooks.

\subsection{Context-sentences}
We extracted LaTeX formulas as well as sentences immediately preceding and following each formula which is refered to as a context-sentence.
This decision considers that in regular discourse, formulas are often discussed with additional expressions preceding and following each formula, rather than in isolation. These context-sentences usually serve to explain or emphasize the formulas. It is also useful for the task of detecting the portions of a spoken sentence that correspond to the formula being mentioned.
For instance, consider a mathematical lecture on YouTube, where viewers see mathematical expressions through real-time subtitles. To effectively present these expressions in LaTeX, it is crucial to quickly identify the formula from preceding texts. Similarly, the text following the formula indicates the end of the formula. Therefore, context-sentences are essential for distinguishing between mathematical and non-mathematical sections of text.

The data structure is as follows. Consider the sentence, ``Here is another equation, \$\$ a\textasciicircum2 + b\textasciicircum2 = c\textasciicircum2 \$\$, which is known as the Pythagorean equation and is very important.''. In the MathBridge dataset, this sentence would be split into `context\_before' column containing ``Here is another equation,'', the `equation' column containing ``\$\$a\textasciicircum2 + b\textasciicircum2 =c\textasciicircum2\$\$'', and `context\_after' column containing ``, which is known as the Pythagorean equation and is very important.''

Extraction was performed simultaneously with LaTeX formula extraction. When extracting formulas using a parser configured for mathematical conditions, the context-senteces were extracted until another formula appeared or a line break occurred. It is worth noting that some data points in MathBridge do not have context-before and/or context-after the formula.

\subsection{Spoken Sentence}
We collected formulas and are now proceeding to obtain the corresponding spoken sentence pairs for each formula using a large language model. Specifically, we employed models such as GPT-3.5, which has 175 billion parameters. By inputting formulas in the LaTeX format into the model and requesting their English vocalization, we verified that these language models produce superior outputs. This capability stems from their extensive pre-training on vast corpora.

We extracted spoken English text using the GPT-3.5 API for the gathered equations. During this process, we only input the formulas, excluding any context-sentence. Considering the presence of duplicate formulas, we temporarily created a dataset that contained unique formulas. This dataset was then used to apply the API, generating spoken English text for approximately 13 million unique formulas out of the initial 49 million (48 million from arXiv and 1 million from textbooks) collected. These results were later aligned with the original data.

For the system role prompt in the GPT-3.5 API, we entered the following text to ensure performance comparable to web-base ChatGPT: You are a ChatGPT, a large language model trained by OpenAI, based on the GPT-3.5 architecture; Knowledge cutoff: 2022-01; Current date: 2024-05-08; Personality: v2.'' For the user role prompt, we provided ``Translate the following LaTeX equation into spoken English without adding any extra explanations:'' along with three examples \cite{few-shot1,few-shot2}. It is well-documented that including examples in the prompt engineering generally enhances performance \cite{promptengineering1,promptengineering2,promptengineering3}. Additionally, we instructed the model to output 'None' if it could not recognize a formula, to prevent the generation of incorrect spoken English text due to potential hallucinations in cases of erroneous formulas.

\begin{figure}[t]
    \centering
    \includegraphics[width=1\linewidth, trim=0mm 7mm 0mm 33mm, clip]{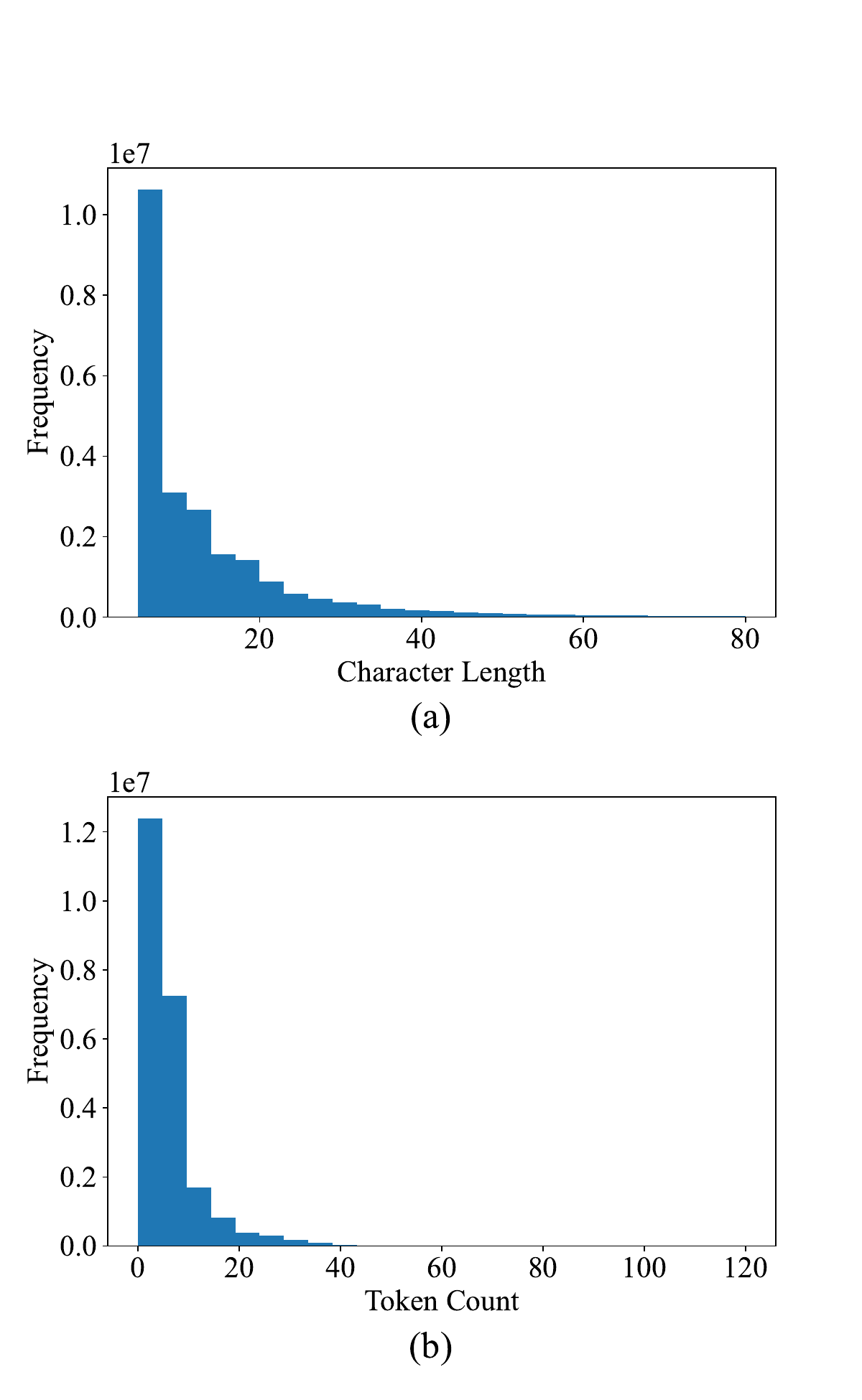}
    \caption{(a) Character length histogram of LaTeX formula data. (b) Token count histogram of spoken sentence.}
    \label{fig:enter-label}
\end{figure}

\subsection{Post-Processing}
For a model to perform well in supervised learning, it is crucial to prepare high-quality data. We rigorously filtered out grammatically incorrect LaTeX data by excluding data points where the GPT-3.5 API generated spoken English text containing terms such as `None', `Sorry', `Apologize', or `cannot', indicating an inability to vocalize the formula.
In addition, any context-sentences containing special characters or TeX commands was completely removed. Upon manually reviewing the dataset, we discovered that data points with excessively long lengths were highly likely to be incorrect. Consequently, we only included data in MathBridge where `context\_before' has no more than 200 characters, `equation' has no more than 80 characters, `context\_after' has no more than 250 characters, and `spoken\_English' has no more than 120 characters. The reason for setting the length of `context\_before' shorter than `context\_after' was based on the length distribution of the data before post-processing, which showed that the sentences preceding the formulas were typically shorter as shown in Figure 4. This reflects the common practice of mentioning the formula first, followed by a more detailed explanation. This process led to the removal of fewer than 500 K data points, maintaining a total of approximately 23 million data points on MathBridge.

\subsection{Dataset Statistics}

In this section, we discuss the statistics of the MathBridge dataset after completing post-processing.
For the formula data, as the histogram in Figure 3(a) shows, the minimum character length was 5, maximum was 80, and the average was 13. Note that there are approximately 7 million data points with the minimum character length. These data points consist of only one character, such as ``\$ x \$'', which includes spaces and dollar signs to make up a length of 5. The prevalence of such data points is substantial because single-character inline formulas are frequently used in academic papers and textbooks. The relevance of these data may vary depending on the specific task the researcher intends to address. For instance, they can be necessary for fine-grained detection tasks that require the identification of single-character formulas. 
\begin{figure}
    \centering
    \includegraphics[width=1\linewidth, trim=10mm 0mm 25mm 12mm, clip]{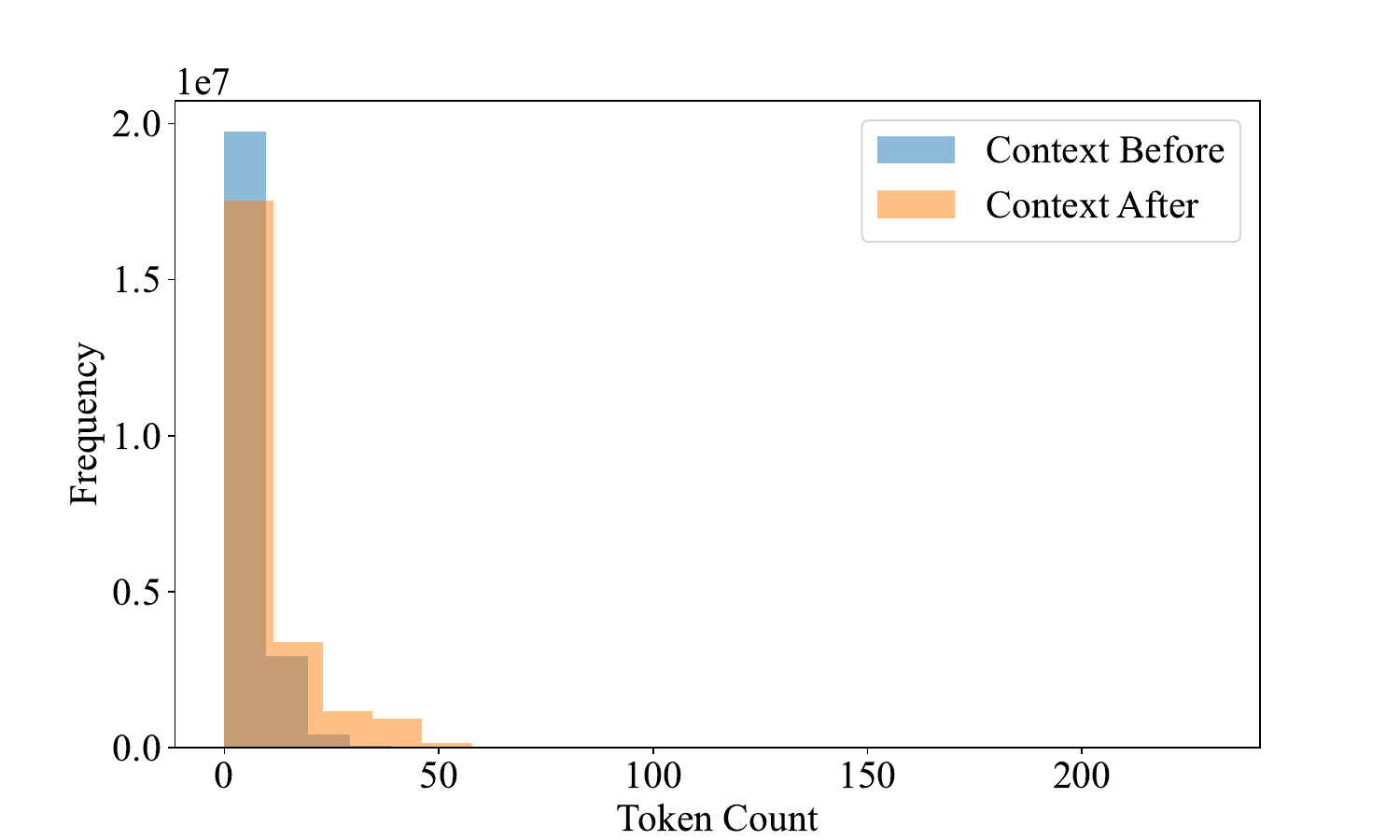}
    \caption{Token count histogram of context-before data and context-after data.}
    \label{fig:enter-label}
\end{figure}

In this analysis, we used the T5 tokenizer \cite{T5} to count the number of tokens in both the context-sentences and spoken English data. For the spoken English data, as the histogram in Figure 3(b) shows, the average token count was 5.79, indicating a distribution similar to that of the formula. This similarity occurs because the lengths of LaTeX formulas and their corresponding spoken English texts are approximately proportional.

For the context-sentences, as illustrated by the histogram in Figure 4, the data for both the before and after texts were biased towards being shorter. However, there was a slight difference in the length distribution between context-before and context-after. The average token count for the before text was 4.89, while the after text had an average of 8.95 tokens, indicating that the context-before is shorter than the after. This difference is due to the tendency to provide more detailed explanations after the formula is mentioned.

\section{Experiment}
We conducted experiments to translate the spoken English of mathematical formulas into the LaTeX format. For this purpose, we utilized the MathBridge dataset to fine-tune various pretrained language models.
\begin{table*}[ht]
\centering
\small 

\begin{tabularx}{\textwidth}{@{}l|*{10}{>{\centering\arraybackslash}X}@{}}

\toprule
\textbf{Models} & \multicolumn{5}{c}{\textbf{Original}} & \multicolumn{5}{c}{\textbf{MathBridge-enhanced}} \\
\cmidrule(lr){2-6} \cmidrule(lr){7-11} & \textbf{BLEU}
 & \textbf{sBLEU} & \textbf{Rouge1} &  \textbf{CER} & \textbf{WER}  & \textbf{BLEU} & \textbf{sBLEU} & \textbf{Rouge1}  & \textbf{CER} & \textbf{WER} \\
 &(\textuparrow) &(\textuparrow) &(\textuparrow) &(\textdownarrow) &(\textdownarrow) &(\textuparrow) &(\textuparrow) &(\textuparrow) &(\textdownarrow) &(\textdownarrow) \\
\midrule
BART-base & 0.29 & 31.3 & 0.64 & 0.51 & 0.68 & 0.26& 38.7& 0.64& 0.42& 0.58   \\
BART-large & 0.29 & 31.0 & 0.61 & 0.52 & 0.69 & 0.31 & 35.2 & 0.63 & 0.48 & 0.54 \\
T5-small & 0.12 & 14.9 & 0.39 & 1.25 & 1.35 & 0.31 &38.4 & 0.75& 0.35& 0.55  \\
T5-base & 0.05 & 6.05 & 0.21 & 2.63 & 2.53 &0.28 &36.6 &0.67 &0.50 &0.74  \\
T5-large &0.04& 4.77 &0.20 & 1.92 & 1.95 &\textbf{0.36} & \textbf{46.8}& \textbf{0.82}& \textbf{0.26}& \textbf{0.49}   \\
mBART-large-50 &   0.21 & 16.9 &0.42 & 0.90&1.37 &0.24 & 23.6 & 0.59 & 0.58 & 0.74  \\
GPT-3.5(w/o p) & 0.24 & 38.9& 0.77& 0.43&0.55  &- &- & -& -& - \\
GPT-3.5(w/ p) & 0.44& 52.3& 0.88 & 0.19 & 0.37&- &- &- & -& - \\
\midrule
\textbf{Average} & 0.16 & 17.4 & 0.41 & 1.28 & 1.42 & 0.29 & 36.5 & 0.68 & 0.43 & 0.60 \\
\midrule
\textbf{Improvement(\%)} & - & - & - & - & - & 76.0 & 109.0 & 65.9 & 66.4 & 57.5 \\
\bottomrule
\end{tabularx}
\caption{Evaluation of the performance of PLMs' original and MathBridge-enhanced responses using the test dataset. Note: `sBLEU' refers to sacreBLEU \cite{sacreBLUE}. The averages for the `Original' column exclude GPT-3.5 Models.}
\end{table*}

\subsection{Dataset}

To enable effective training for translating from English to LaTeX, we excluded approximately 7 million data points, each consisting of only one character, from both the training and testing datasets within the MathBridge collection. From the remaining dataset of approximately 16 million entries, we initially selected 1000 pairs of LaTeX formulas and their corresponding spoken English expressions for the testing dataset. For this dataset, denoted by $D_{test}={(x_i,y_i )}_{i=1}^{N}$, each input $x_i$ was constructed by concatenating the context\_before, the spoken English version of the formula, and the context\_after. Similarly, each target output, $y_i$, is formed by concatenating context\_before, the LaTeX formula, and context\_after. The remaining data were used as the training set.
Following the same structure for the training dataset, denoted as $D_{train}={(x_i,y_i )}_{i=1}^{N}$, each input, $x_i$, and output, $y_i$, were assembled in the same manner by concatenating the respective texts before and after the formula, with either the spoken English or the LaTeX formula in between. This approach ensured a structured and consistent framework for training our models to understand and translate the contextual relationships between spoken English and LaTeX formulas effectively.

\begin{table}[]
    \centering
    \setlength{\tabcolsep}{1pt}
    \footnotesize
    \begin{tabular}{>{\raggedright\arraybackslash}p{2.5cm} >{\raggedleft\arraybackslash}p{2.5cm}}
    \toprule
        \textbf{Model} & \textbf{Parameters} \\
    \midrule
        BART-base & 139 M\\
        BART-large & 406 M\\
        T5-small & 60.5 M\\
        T5-base & 223 M\\
        T5-large & 738 M\\
        mBART-large-50 & 406 M\\
        GPT-3.5 & 175 B\\
    \bottomrule
    \end{tabular}
    \caption{Parameter sizes of the models used in the experiment.}
    \label{tab:my_label}
\end{table}
\subsection{Setup}
We selected several SOTA PLMs, such as T5 \cite{T5}, BART \cite{BART}, and mBART \cite{mBART}, which have demonstrated satisfactory performance on other downstream tasks, as baselines for translating English to LaTeX. The maximum number of training epochs was set to 5, and the model that achieved the lowest validation loss on the development set was selected as the best model for prediction on the test set. The learning rate was adjusted using a cosine learning rate scheduler, with the maximum learning rate set at 1e-4 and the minimum at 1e-6. The maximum input sequence length was set to 512, and the batch size was 64. We used NVIDIA A100.

For some current powerful large-language models, such as GPT-3.5, which incur significant computational costs, we evaluated their translation capabilities solely through inference. We also compared the performance differences between when prompts were provided and when they were not \cite{talkfunny}.  The prompt instruction used was, ``The following sentence mixes spoken parts of formulas with English. Translate the part of the sentence that represents a formula into LaTeX.''. This prompt demonstrated the best performance in preliminary experiments.

\subsection{Metrics}
To evaluate the translated LaTeX format, we employed traditional metrics commonly used in translation tasks. These metrics include sacreBLEU \cite{sacreBLUE,BLEU}, ROUGE \cite{ROUGE}, CER, and WER. SacreBLEU, a variation of BLEU, offers high reliability because it produces consistent values regardless of the tokenizer used. It is also applicable not only to English but also to other languages. Given that LaTeX often has special characteristics, we included sacreBLEU in our metrics suite. In addition, we utilized ROUGE, a metric typically employed in summarization tasks, which is a subset of translation tasks. ROUGE was used to gauge the quality of textual summaries. In Table 3, Rouge1 represents the precision of the ROUGE-1 Mid. Furthermore, we chose the traditional language accuracy metrics, WER and CER, to assess language alignment.

\begin{figure*}[t]
    \centering
    \includegraphics[width=1\textwidth, trim=27mm 80mm 27mm 80mm, clip]{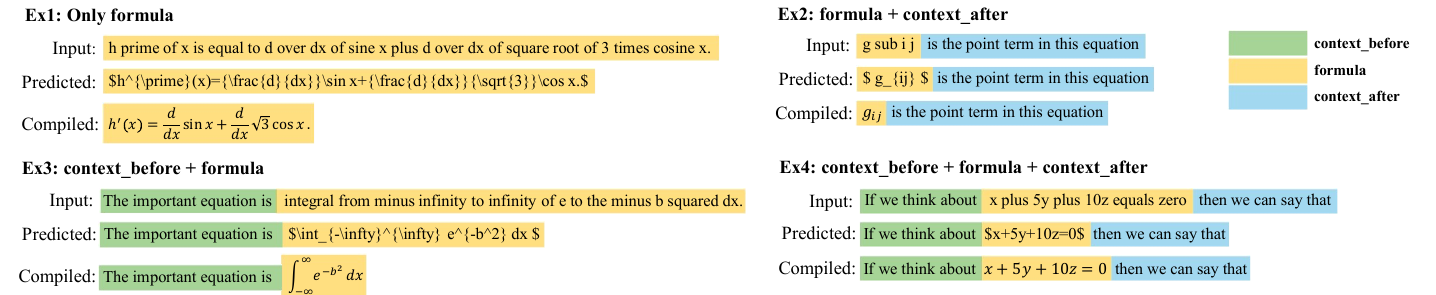}
    \caption{Four good cases generated by fine-tuned PLMs based on MathBridge.}
    \label{fig:enter-label}
\end{figure*}

\subsection{Results}
The results are presented in Table 3. The initial performance of the PLMs on the test dataset was found to be unsatisfactory, indicating the limited capability of these models to convert spoken mathematical sentences into LaTeX without specific adaptations. Notable performance enhancements were observed after fine-tuning the PLMs using MathBridge, particularly for the T5-large model. Furthermore, as shown in Table 4, although GPT-3.5 is a large model with 175 billion parameters, the T5-large model has 237 times fewer parameters than GPT-3.5. Despite its smaller size, it has shown results that exceed those of GPT-3.5 without prompts. These improvements indicate MathBridge's efficacy and suitability for translating English into LaTeX. Because LaTeX is compatible with markup languages, it can be easily compiled and will be useful.

Moreover, the effectiveness of advanced PLMs, such as GPT-3.5, without prompt instructions, was inadequate for this conversion task. However, significant improvements were achieved in the test dataset after integrating prompt instructions with GPT-3.5, highlighting the critical role of prompt engineering in enhancing the performance of PLMs for this specific application.

\subsection{Case Study and Error Analysis}

In this section, we present several good examples generated using our fine-tuned PLMs. We also analyzed some problematic cases. Example 1 in Figure 5 illustrates that the PLM, fine-tuned with MathBridge, robustly predicts LaTeX, even for relatively complex equations. Moreover, in Example 4 of the same figure, the input spoken English was, ``If we think about x plus 5y plus 10z equals zero then we can say that." The fine-tuned PLMs with our dataset effectively distinguished the mathematical parts and accurately predicted the LaTeX equations. These equations can subsequently be compiled into formulas using a LaTeX compiler, commonly rendering all formulas as seen in the compiled outputs.

However, we identified challenges related to the interpretation of spoken English formulas. A single spoken expression corresponds to multiple mathematical representations. For example, consider the input: ``F of x equals the square root of 10 times x to the power of negative 8 divided by negative 8, plus C sub 1, equals negative 5 divided by 4 times x to the power of 8 plus C sub 1, if x is less than 0." The compiled formula from the predicted LaTeX code using the fine-tuned T5-large model was as follows:

\begin{equation}
F(x)=\sqrt{10}x^{-8/-8}+C_{1}=-\frac{5} {4}x^{8}+C_{1}\quad\mathrm{if~}x<0.
\end{equation}

However, when the ground truth LaTeX from the dataset was compiled, it resulted in:

\begin{equation}
F(x)={\sqrt{\frac{10x^{-8}}{-8}}}+C_{1} =-{\frac{5}{4x^{8}}}+C_{1}\quad{\mathrm {if~}}x<0.
\end{equation}

This example highlights the ambiguity that a single spoken formula expression can imply multiple mathematical interpretations. A closer look reveals that ambiguity primarily stems from the unclear boundaries of the phrase ``the square root of'' which the model appears to have learned based on the presence of a comma. Addressing this ambiguity may require a more context-aware interpretation or more precise expressions in spoken input.

\section{Discussion}
LaTeX is a language comprised of a set of commands that include special characters, more similar to computer languages such as C and SQL than to general natural languages such as English. The commonly used BLEU metric \cite{BLEU} for translation tasks is based on a tokenizer. This tokenizer segments sentences based on a general natural language. However, it is unclear whether the traditional tokenizer can encapsulate information in LaTeX expressions, making BLEU an imperfect metric for evaluating LaTeX. Although sacreBLEU offers a more robust evaluation than BLEU, it still relies on the congruence of n-grams and fails to recognize semantic equivalence. Also, the `\textbackslash frac' command and a simple `/' both denote fractions, yet this semantic identity is not reflected in its assessments. Character Error Rate (CER) shares similar issues with sacreBLEU. The Word Error Rate (WER), which separates words by spaces and calculates errors on a word basis, also does not serve as an ideal metric for LaTeX, where spaces do not always indicate syntactic separation. Thus, an ideal metric for assessing the LaTeX format should meet the following criteria:
\begin{itemize}
\item LaTeX expressions that compile into the same formulas should be evaluated as identical.
\item The metric should quantify LaTeX error rates without being affected by spacing.
\item LaTeX expressions, such as equation 1 and equation 2, which might look similar but carry completely different meanings, should be distinctly assessed.
\end{itemize}
We hope to continuously expand our applications based on our dataset in the future.

\section{Conclusion and Future Work}
Translating spoken mathematical expressions into LaTeX format is a challenging task in NLP. In this study, we introduced MathBridge, the first large-corpus dataset. Utilizing this dataset, we extensively evaluated the translation abilities of PLMs and enhanced their performance in converting spoken English into the LaTeX format. In addition, we analyzed the performance differences between large language models with and without prompts using our test dataset. The experimental results demonstrate that our method can effectively assist PLMs in generating LaTeX. In the future, researchers can explore swapping the input and ground-truth data used for fine-tuning, thereby enabling the conversion of the LaTeX format to spoken expression. This could lead to the development of a specialized TTS system for mathematical formulas.  In addition, they can create a more general and robust TeX tokenizer and metric to evaluate LaTeX text.

\bibliography{references}

\end{document}